\pdfoutput=1
\documentclass[11pt]{article}

\usepackage[final]{acl}

\usepackage{times}
\usepackage{latexsym}

\usepackage[T1]{fontenc}
\usepackage[utf8]{inputenc}

\usepackage{microtype}

\usepackage{inconsolata}

\usepackage{graphicx}

\usepackage{graphicx}
\usepackage{multirow}
\usepackage{arydshln}
\newcommand{\bhline}{\noalign{\hrule height 1.2pt}}

\title{CiMaTe: Citation Count Prediction\\Effectively Leveraging the Main Text}

\author{Jun Hirako \hspace{25pt} Ryohei Sasano \hspace{25pt} Koichi Takeda \\
    Graduate School of Informatics, Nagoya University \\
    \texttt{junhirako17@gmail.com} \\
    \texttt{\{sasano,takedasu\}@i.nagoya-u.ac.jp} \\
}

\begin{document}
\maketitle
\begin{abstract}
Prediction of the future citation counts of papers is increasingly important to find interesting papers among an ever-growing number of papers.
Although a paper's main text is an important factor for citation count prediction, it is difficult to handle in machine learning models because the main text is typically very long; thus previous studies have not fully explored how to leverage it.
In this paper, we propose a BERT-based citation count prediction model, called CiMaTe, that leverages the main text by explicitly capturing a paper's sectional structure.
Through experiments with papers from computational linguistics and biology domains, we demonstrate the CiMaTe's effectiveness, outperforming the previous methods in Spearman’s rank correlation coefficient; 5.1 points in the computational linguistics domain and 1.8 points in the biology domain.
\end{abstract}

\section{Introduction}

%The number of academic papers has increased dramatically in various fields, notably artificial intelligence.
%Accordingly, techniques to predict the quality of papers are becoming increasingly important for finding interesting papers.
%In this paper, we adopt the citation count as an approximate indicator of a paper's quality~\cite{Chubin2005IsCA,Aksnes2006CitationRA}, and we tackle the task of predicting paper's future citation counts.
The number of academic papers has increased dramatically in various fields.
Accordingly, techniques to automatically estimate the quality of papers are becoming increasingly important for finding intersting papers among the vast number of papers.
In this paper, we adopt the citation count as an approximate indicator of a paper's quality, following previous studies~\cite{Chubin2005IsCA,Aksnes2006CitationRA}, and we tackle the task of predicting paper's future citation counts.

In predicting citation counts, the use of a paper's main text has the potential to improve performance.
%However, most existing studies \citep[e.g.,][]{Ibez2009PredictingCC,Ma2021ADB}) use only titles and abstracts as the textual information.
%A major reason is that a paper's main text is very long and difficult to process.
However, most existing studies \citep[e.g.,][]{Ibez2009PredictingCC,Ma2021ADB} use only titles and abstracts as the textual information.
A major reason for this is that the papers' main texts are very long and therefore difficult to handle in machine learning models.
In particular, the quadratic computational costs of Transformer~\cite{NIPS2017_3f5ee243}-based models such as BERT~\cite{devlin-etal-2019-bert} make it difficult to process long documents.
Approaches to overcome this problem include improvement in the Transformer structure \citep[e.g.,][]{Kitaev2020ReformerTE,Beltagy2020LongformerTL} and processing of long documents by dividing them into several parts \citep[e.g.,][]{Pappagari2019HierarchicalTF,Afkanpour2022BERTFL}, but it is unclear whether these approaches are effective or not in citation count prediction.

In this paper, we propose CiMaTe, a simple and strong \textbf{Ci}tation count prediction model that leverages the \textbf{Ma}in \textbf{Te}xt.
CiMaTe explicitly captures a paper's sectional structure by encoding each section with BERT and then aggregating the encoded section representations to predict the citation count.
We demonstrate CiMaTe's effectiveness by performing comparative experiments with several methods on two sets of papers collected from arXiv and bioRxiv, respectively.
%We demonstrate CiMaTe's effectiveness through extensive comparison experiments with multiple methods on two sets of papers collected from arXiv and bioRxiv, respectively.

\section{Related Work}

\begin{figure*}[t!]
  \centering
  \includegraphics[width=\linewidth]{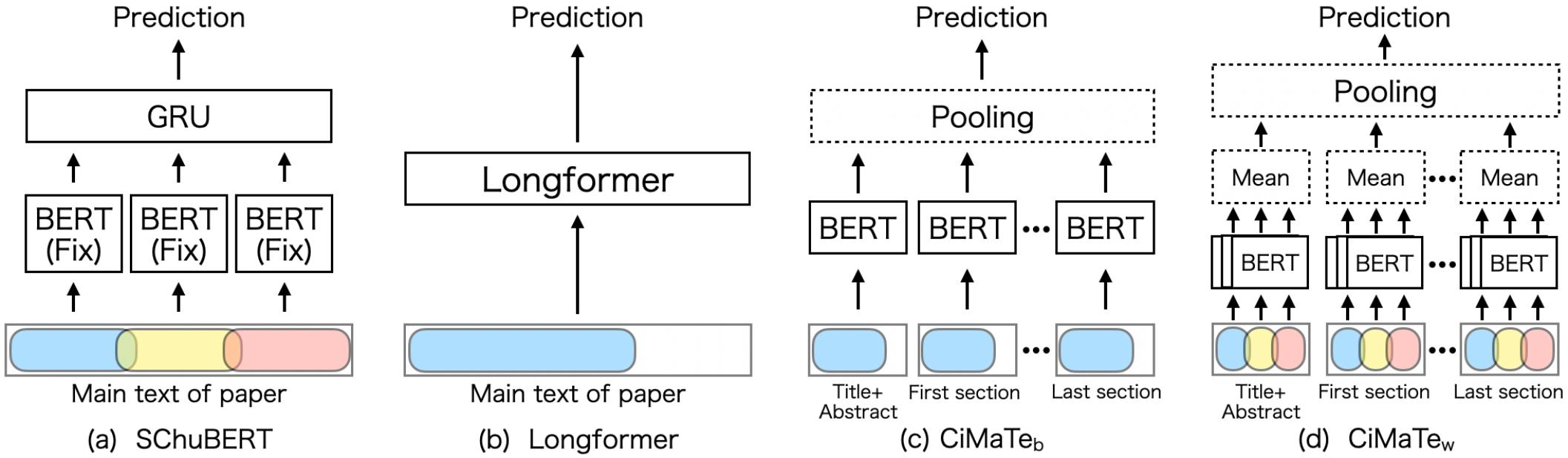} \vspace{-3.2ex}
  \caption{Overview of four citation count prediction methods that leverage a paper's main text.}
  \label{FIG::overviews}
  \vspace{-0.7ex}
\end{figure*}

\subsection{Citation Count Prediction}

Existing methods for predicting citation counts include those that use textual information and those that do not.
Among methods that do not use textual information, those of \citet{Castillo2007EstimatingNO}, \citet{davletov-high-2014}, and \citet{Pobiedina2015CitationCP} represent the citation relationships of papers as a graph. \citet{Abrishami2018PredictingCC} proposed a method that predicts the long-term citation count from short-term citation counts by using an RNN.

As for methods that use textual information, \citet{Ibez2009PredictingCC} proposed a method that uses a bag-of-words with abstracts, \citet{Fu2008ModelsFP} proposed a method that uses frequency-based weighting, and \citet{Yan2011CitationCP} and \citet{Chakraborty2014TowardsAS} proposed methods that use topics estimated by LDA.
In recent years, there has been research on methods to capture the semantic features of papers' textual information through deep learning.
\citet{Ma2021ADB} proposed a method that encodes the title and abstract with Doc2Vec \cite{DBLP:conf/icml/LeM14}, and \citet{Dongen2020SChuBERTSD} proposed a method that uses a paper's main text via a pre-trained model.
\citet{hirako-etal-2023-realistic} proposed a BERT-based method that captures the latest research trends for newly-published papers.

\subsection{Long Document Processing}

There are two major approaches to processing long documents in a Transformer-based model.
The first approach modifies the Transformer structure to efficiently process long sequences~ \cite{Kitaev2020ReformerTE,Beltagy2020LongformerTL,Zaheer2020BigBT}.
These methods mainly modify the self-attention module and improve the computational complexity.
The second approach processes long documents by dividing them up.
This approach includes methods that encode the divided text with BERT and aggregate it with simple techniques such as attention or an RNN~\cite{Afkanpour2022BERTFL,Dongen2020SChuBERTSD}.
The second approach also includes hierarchical methods in which a BERT-encoded representation is further input to a Transformer~\cite{Pappagari2019HierarchicalTF,9530411}.

\section{Leveraging the Main Text}
\label{SEC::prediction_model}

\subsection{Existing Approaches}

\textbf{SChuBERT}~\cite{Dongen2020SChuBERTSD} is an existing method for predicting citation counts from the whole main text of papers.
As shown in Figure \ref{FIG::overviews} (a), it predicts the citation count by dividing the main text into overlapping chunks, encoding each of them with BERT, and aggregating them with a GRU~\cite{cho-etal-2014-learning}.
As SChuBERT does not fine-tune BERT, it can leverage the whole main text with a relatively small computational cost.
However, it divides the main text with a fixed number of characters and does not explicitly capture a paper's structure.

\textbf{Longformer}~\cite{Beltagy2020LongformerTL}, a pre-trained language model for processing long sequences, can be used to predict citation counts leveraging the main text.
As shown in Figure \ref{FIG::overviews} (b), this method inputs the beginning of the main text by truncating it to a maximum input length of 4096 tokens.
While this truncation is commonly used~\cite{NEURIPS2020_44feb009}, much of a paper's main text may be truncated, and Longformer thus cannot consider the whole paper equally.
For example, it truncated about 46\% of the main text on average for the dataset used in our experiments (see Section \ref{SEC::dataset}).

\subsection{CiMaTe}

Our proposed method, CiMaTe predicts citation counts by encoding each section with BERT and then pooling the encoded section representations.
We introduce two versions of CiMaTe: \textbf{CiMaTe$_{\mbox{\small b}}$}, which leverages only the \textbf{b}eginning of each section, and \textbf{CiMaTe$_{\mbox{\small w}}$}, which divides each section to leverage the \textbf{w}hole main text.
Figure \ref{FIG::overviews} (c) shows an overview of CiMaTe$_{\mbox{\small b}}$.
This method inputs the section title as the first sentence and the content as the second sentence into BERT to generate section representations.
The input content is truncated to not exceed 512 tokens, BERT's maximum input length.
While this method mitigates the increase in computational cost, only limited content is taken into account for each section.

Figure \ref{FIG::overviews} (d) shows an overview of CiMaTe$_{\mbox{\small w}}$.
First, this method divides the content of each section into chunks; then it encodes each chunk with BERT.
The chunk representations are then pooled with their means to generate section representations.
Following previous studies~\cite{Pappagari2019HierarchicalTF,Dongen2020SChuBERTSD}, each content is divided into chunks that overlap by 50 tokens each.
In encoding each chunk, the section title is input as the first sentence, and the chunk, as the second sentence, even if it is the second or later chunk.
While this method can take the whole content into account, it increases the computational cost.

These two methods predict citation counts by pooling section representations.
For a more expressive pooling method, we use Transformers besides the means.
In Transformer pooling, each section representation is input to a one-layer Transformer as a sequence, and then the output is averaged.

\section{Experiments}

\subsection{Task Setting}

To validate the effectiveness of the CiMaTe, we compared several methods on the realistic citation count prediction task for newly published papers proposed by \citet{hirako-etal-2023-realistic}.
In this task, the information used for training is strictly restricted to that available at the time of the target paper's publication, which allows for a more realistic evaluation of the prediction method.
The target of the prediction is the citation counts one year after publication.
While the citation count one year after publication is available for papers published more than one year before the target paper, it is not available for papers published less than one year before the target paper.
Thus, we used the method proposed by \citet{hirako-etal-2023-realistic} to complement the citation counts for these papers.
The specific target of the prediction is the logarithm $\mbox{log}(c+1)$ rather than the citation count $c$, reflecting the intuition that the difference between 0 and 10 cases is more significant than between 1,000 and 1,010 cases.

\subsection{Dataset}
\label{SEC::dataset}

As with \citet{hirako-etal-2023-realistic}, we constructed two datasets from the preprint server papers: a CL dataset from papers in the computational linguistics field, and a Bio dataset from the biology field.
The reason for using papers from the preprint server was that while a large number of papers are published, some papers are not peer-reviewed and their quality is not guaranteed, so it was considered highly important to predict citation counts.

The CL dataset was constructed from papers submitted to the cs.CL category of arXiv\footnote{\url{https://arxiv.org/}} between June 2014 and June 2020.
Citation information for calculating the citation counts was collected from Semantic Scholar.\footnote{\url{https://www.semanticscholar.org/}}
The main texts of papers were collected by parsing their HTML versions.\footnote{\url{https://ar5iv.labs.arxiv.org/}}
Papers for which HTML versions could not be collected and parsed were excluded from the dataset.
We created 13 subsets of the CL dataset.
Specifically, starting with each month between June 2019 and June 2020 inclusive, we created 13 subsets comprising papers that had been submitted in the previous 5 years before each of those months.
The papers submitted in the last month of each subset were used for evaluation, while the other papers in the subset were used for training.
Among these subsets, the one using papers published in June 2019 for evaluation was used as the development set, while the remaining subsets were used to calculate the evaluation scores.
The average numbers of papers for training and evaluation were 11,383 and 449, respectively.

The Bio dataset was constructed from papers submitted to the Biochemistry, Plant Biology, and Pharmacology and Toxicology categories of bioRxiv\footnote{\url{https://www.biorxiv.org/}} between April 2015 and April 2021.
The main texts were collected by parsing the papers' HTML versions on bioRxiv.
As with the CL dataset, we created 13 subsets of the Bio dataset.
Papers submitted in each month from April 2020 to April 2021 inclusive were used for evaluation, and the subset with the oldest papers for evaluation was again used as the development set.
The average numbers of papers for training and evaluation were 5,737 and 256, respectively.

\subsection{Evaluation}

For evaluation metrics, we used Spearman's rank correlation coefficient ($\rho$), the mean squared error (MSE), and a metric defined as the percentage of the actual top n\% of papers in the top k\% of the output (n\%@k\%).
We calculated the evaluation scores as follows.
First, for each subset, we trained a model and use that model to predict the citation count for each paper.
We then compiled the predictions of citation counts for the 12 subsets and calculated an evaluation score on the whole.
In all settings, we experimented with three different random seeds and calculated the means and standard deviations of the evaluation scores.

For certain methods, we found in our experiments that the predictions tend to be slightly larger than the correct values, which had a negative impact on the appropriate evaluation using MSE.
To mitigate this, we also used MSE*, a variant of the MSE, as an evaluation metrics.
This metric calculates the MSE by subtracting the difference between the average of the predictions and the average of the correct values for the development set from the predictions at test time.

\addtolength{\tabcolsep}{-3.5pt}
\begin{table*}[t!]
    \small
    \centering
    \begin{tabular}{@{}cl@{}lrrrrrrr}
    \bhline
    \multicolumn{1}{c}{Dataset} & \multicolumn{1}{c}{Method} & \multicolumn{1}{c}{Pooling} & \multicolumn{1}{c}{$\rho$} & \multicolumn{1}{c}{MSE} & \multicolumn{1}{c}{MSE*} & \multicolumn{1}{c}{5\%@5\%} & \multicolumn{1}{c}{5\%@25\%} & \multicolumn{1}{c}{10\%@10\%} & \multicolumn{1}{c}{10\%@50\%} \\ \hline
    \multirow{8}{*}{CL} & BERT${}_{\scriptsize \mbox{t+a}}$ & \multicolumn{1}{c}{-} & 37.0${}_{\pm 0.2}$ & 1.295${}_{\pm .016}$ & 1.082${}_{\pm .014}$ & 28.8${}_{\pm 0.8}$ & 69.2${}_{\pm 1.0}$ & 34.1${}_{\pm 1.9}$ & 83.9${}_{\pm 0.4}$ \\
     & BERT${}_{\scriptsize \mbox{b}}$ & \multicolumn{1}{c}{-} & 38.5${}_{\pm 0.8}$ & 1.246${}_{\pm .047}$ & 1.028${}_{\pm .010}$ & 29.0${}_{\pm 2.1}$ & 71.5${}_{\pm 1.5}$ & 34.6${}_{\pm 1.7}$ & 86.0${}_{\pm 1.0}$ \\
     & SChuBERT & \multicolumn{1}{c}{-} & 37.5${}_{\pm 0.9}$ & \textbf{0.983}${}_{\pm .015}$ & 0.987${}_{\pm .014}$ & 24.7${}_{\pm 3.3}$ & 62.3${}_{\pm 2.1}$ & 29.0${}_{\pm 1.4}$ & 82.2${}_{\pm 1.3}$ \\
     & Longformer & \multicolumn{1}{c}{-} & 37.9${}_{\pm 0.1}$ & 1.245${}_{\pm .027}$ & 1.046${}_{\pm .014}$ & 27.9${}_{\pm 1.4}$ & 74.3${}_{\pm 1.6}$ & 34.6${}_{\pm 0.3}$ & 85.8${}_{\pm 0.3}$ \\
     & CiMaTe${}_{\scriptsize \mbox{b}}$ & mean & 42.6${}_{\pm 0.1}$ & 1.184${}_{\pm .025}$ & \textbf{0.955}${}_{\pm .006}$ & 33.1${}_{\pm 1.7}$ & 74.0${}_{\pm 2.0}$ & 38.3${}_{\pm 0.4}$ & 86.7${}_{\pm 0.9}$ \\
     & CiMaTe${}_{\scriptsize \mbox{b}}$ & Transformer & \textbf{43.6}${}_{\pm 0.6}$ & 1.228${}_{\pm .051}$ & 0.976${}_{\pm .015}$ & 33.6${}_{\pm 0.4}$ & \textbf{76.7}${}_{\pm 1.3}$ & \textbf{38.4}${}_{\pm 0.7}$ & \textbf{88.1}${}_{\pm 1.1}$ \\
     & CiMaTe${}_{\scriptsize \mbox{w}}$ & mean & 42.9${}_{\pm 0.5}$ & 1.157${}_{\pm .005}$ & 0.958${}_{\pm .009}$ & 32.7${}_{\pm 1.1}$ & 75.4${}_{\pm 0.6}$ & 37.3${}_{\pm 0.5}$ & 87.3${}_{\pm 0.9}$ \\
     & CiMaTe${}_{\scriptsize \mbox{w}}$ & Transformer & 43.2${}_{\pm 1.2}$ & 1.170${}_{\pm .013}$ & 0.985${}_{\pm .019}$ & \textbf{34.2}${}_{\pm 0.9}$ & 74.0${}_{\pm 0.8}$ & 37.8${}_{\pm 1.7}$ & 87.8${}_{\pm 1.0}$ \\ \hline \hline
    \multirow{8}{*}{Bio} & BERT${}_{\scriptsize \mbox{t+a}}$ & \multicolumn{1}{c}{-} & 34.7${}_{\pm 0.5}$ & 0.517${}_{\pm .012}$ & 0.518${}_{\pm .012}$ & 48.8${}_{\pm 0.4}$ & 82.1${}_{\pm 0.8}$ & 47.4${}_{\pm 1.3}$ & 86.0${}_{\pm 1.0}$ \\
     & BERT${}_{\scriptsize \mbox{b}}$ & \multicolumn{1}{c}{-} & 38.4${}_{\pm 0.7}$ & 0.484${}_{\pm .010}$ & 0.489${}_{\pm .009}$ & 49.5${}_{\pm 1.0}$ & 85.8${}_{\pm 2.3}$ & \textbf{51.5}${}_{\pm .0}$ & 87.8${}_{\pm 2.1}$ \\
     & SChuBERT & \multicolumn{1}{c}{-} & 27.2${}_{\pm 4.8}$ & 0.663${}_{\pm .015}$ & 0.613${}_{\pm .023}$ & 32.2${}_{\pm 2.5}$ & 65.6${}_{\pm 2.3}$ & 35.5${}_{\pm 2.3}$ & 74.4${}_{\pm 3.5}$ \\
     & Longformer & \multicolumn{1}{c}{-} & 39.1${}_{\pm 0.2}$ & 0.479${}_{\pm .005}$ & 0.484${}_{\pm .007}$ & 50.3${}_{\pm 1.7}$ & 84.7${}_{\pm 2.6}$ & 49.6${}_{\pm 0.2}$ & 86.4${}_{\pm 0.7}$ \\
     & CiMaTe${}_{\scriptsize \mbox{b}}$ & mean & 39.4${}_{\pm 0.7}$ & \textbf{0.450}${}_{\pm .008}$ & \textbf{0.448}${}_{\pm .008}$ & 51.4${}_{\pm 2.7}$ & 86.5${}_{\pm 0.4}$ & 51.4${}_{\pm 0.7}$ & 90.2${}_{\pm 1.2}$ \\
     & CiMaTe${}_{\scriptsize \mbox{b}}$ & Transformer & \textbf{40.9}${}_{\pm 0.5}$ & 0.452${}_{\pm .006}$ & 0.453${}_{\pm .007}$ & \textbf{52.7}${}_{\pm 1.0}$ & 87.8${}_{\pm 1.0}$ & 51.4${}_{\pm 0.5}$ & 90.8${}_{\pm 0.2}$ \\
     & CiMaTe${}_{\scriptsize \mbox{w}}$ & mean & 38.5${}_{\pm 0.0}$ & 0.467${}_{\pm .006}$ & 0.461${}_{\pm .009}$ & 51.0${}_{\pm 1.3}$ & 86.7${}_{\pm 1.0}$ & 49.6${}_{\pm 0.7}$ & 88.2${}_{\pm 1.4}$ \\
     & CiMaTe${}_{\scriptsize \mbox{w}}$ & Transformer & 40.7${}_{\pm 0.9}$ & 0.456${}_{\pm .001}$ & 0.457${}_{\pm .001}$ & 52.1${}_{\pm 1.5}$ & \textbf{89.1}${}_{\pm 1.5}$ & 50.3${}_{\pm 0.8}$ & \textbf{91.0}${}_{\pm 1.0}$ \\ \bhline
    \end{tabular}
    \caption{Experimental results for comparison of citation count prediction methods. Besides the MSE and MSE*, each score is multiplied by 100.}
    \label{TAB::main_results}
\end{table*}
\addtolength{\tabcolsep}{3.5pt}

\subsection{Comparison Methods}

In addition to the methods described in Section \ref{SEC::prediction_model}, we compared two baseline methods using BERT.
First, BERT${}_{\small \mbox{t+a}}$ was a method to predict citation counts by inputting the \textbf{t}itle as the first sentence and the \textbf{a}bstract as the second sentence into BERT.
Second, BERT$_{\small {\mbox{b}}}$, was a method to predict citation counts by inputting the \textbf{b}eginning of the main text into BERT, similarly to Longformer, but this method used only a maximum of 512 tokens, whereas Longformer allows a maximum of 4096 tokens to be input.

\subsection{Experimental Settings}

% SChuBERT was trained in the same setting as in a previous study \cite{Dongen2020SChuBERTSD}, except for the range of hyperparameter search.
We used the vector representation corresponding to the \verb|[CLS]| token as the output of BERT and Longformer; then, they input the model's final vector output to a fully connected layer to predict citation counts.
In CiMaTe$_{\mbox{\small w}}$, to somewhat mitigate the computational cost, the content of each section was divided into a maximum of eight chunks, and the remaining text was truncated.
The loss function was the MSE of the predictions and the correct values, and we applied dropout~\cite{JMLR:v15:srivastava14a} to the final output vector during training.

For pre-trained weights, we used `bert-base-uncased' for BERT and `allenai/longformer-base-4096' for Longformer, both of which are publicly available on the Transformers~\cite{wolf-etal-2020-transformers}.
Also, there is no Longformer variant that was pre-trained on the scientific domain corpus.
To make a fair comparison, we used only BERT that was pre-trained on the general domain corpus and did not use models pre-trained on the scientific domain corpus, such as SciBERT~\cite{beltagy-etal-2019-scibert}.

All methods except SChuBERT were trained with a batch size of 32, the AdamW~\cite{loshchilov2018decoupled} optimizer, and a learning rate schedule that warm-up to 10\% of the total training steps and then decayed linearly over the remaining steps.
For the number of epochs and the learning rate, we performed grid search with values of \{3, 4\} for the number of epochs and \{2e-5, 3e-5, 5e-5\} for the learning rate, and we then used the values that yielded the highest rank correlation coefficient on the development set.
For SChuBERT, only the number of epochs was determined by hyperparameter search, while the remaining settings were trained as in previous studies.
The search range for the number of epochs was set to \{20, 30, 40\} according to the values in previous studies.
% \footnote{The details of the settings are provided in Appendix \ref{ADX::training_settings}.}

\subsection{Results}

The experimental results are summarized in Table \ref{TAB::main_results}.
Except for SChuBERT, the methods that leverage a paper's main text outperformed BERT$_{\small \mbox{t+a}}$ on both datasets; thus we confirmed the effectiveness of leveraging the main text in predicting citation counts.
Among the methods that leverage the main text, CiMaTe, which explicitly captures a paper's sectional structure, achieved the highest performance.
Specifically, in terms of the rank correlation coefficient, the model that scored best among the CiMaTe variants outperformed the other methods by at least 5.1 points on the CL dataset and at least 1.8 points on the Bio dataset.
In terms of the MSE, MSE*, and n\%@k\%, it also outperformed the other methods in most cases, which demonstrates the general effectiveness of CiMaTe on sets of papers from different fields, namely, computational linguistics and biology.

Between the two CiMaTe versions, CiMaTe$_{\mbox{\small b}}$ achieved the same or better performance as CiMaTe$_{\mbox{\small w}}$, and we thus found that by using only the beginning of each section, it was possible to achieve high prediction performance while mitigating the increase in computational cost.
We also confirmed that Transformer pooling performed better overall than using means in the pooling of section representations.
Although SChuBERT showed an improved MSE and MSE* on the CL dataset, its performance was inferior to that of BERT$_{\small \mbox{t+a}}$ in other cases.
We speculate that this was because SChuBERT did not fine-tune BERT to accept the whole main text as input with a relatively small computational cost~\cite{Pappagari2019HierarchicalTF}.

\section{Conclusion}

We have proposed CiMaTe, a citation count prediction method that effectively leverages a paper's main text, and demonstrated its effectiveness through experimental comparison with several other methods on sets of papers from different fields, namely, computational linguistics and biology.
In particular, we confirmed that CiMaTe can achieve high performance while mitigating the increase in computational cost by using only the beginning of each section.
For future work, we plan to build a model that simultaneously considers information besides a paper's text, such as the information available from the figures and tables, author information, and citation graphs.

\bibliography{custom}

\end{document}